\documentclass[a4]{article} % For LaTeX2e
\usepackage{times}

\usepackage{hyperref}
\usepackage{url}
\usepackage{eulervm}
\usepackage{charter}
\usepackage[scaled]{beramono}
\usepackage[T1]{fontenc}
\usepackage{graphicx}

\usepackage{amsmath}
\usepackage{amssymb}

\usepackage{float}

\newcommand{\bc}{\mathbold{c}}
\newcommand{\bu}{\mathbold{u}}
\newcommand{\bl}{\mathbold{l}}
\newcommand{\bo}{\mathbold{o}}
\newcommand{\x}{\mathbold{x}}
\newcommand{\bd}{\mathbold{d}}
\newcommand{\bt}{\mathbold{t}}
\newcommand{\br}{\mathbold{r}}
\newcommand{\X}{\mathbold{X}}
\newcommand{\y}{\mathbold{y}}

\newcommand{\bv}{\mathbold{v}}
\newcommand{\D}{\mathbold{D}}
\newcommand{\W}{\mathbold{W}}

\newcommand{\I}{\mathbold{I}}

\newcommand{\sig}{\mathbold{\sigma}}

\newcommand{\R}{{\mathbb R}}

\newcommand{\N}{{\mathbb N}}
\newcommand{\p}{{\;\text{.}}}
\newcommand{\tab}{\hspace*{5mm}}

\floatstyle{ruled}
\newfloat{pseudo}{h}{lop}
\floatname{pseudo}{Algorithm}

\title{Learning sparse transformations through backpropagation}

\author{Peter Bloem \\
Knowledge Representation and Reasoning Group\\
Vrije Universiteit Amsterdam\\
 \\
\texttt{vu@peterbloem.nl} \\
}

\begin{document}

\maketitle

\begin{abstract}

\noindent Many transformations in deep learning architectures are sparsely connected. When such transformations cannot be designed by hand, they can be learned, even through plain backpropagation, for instance in attention mechanisms. However, during learning, such sparse structures are often represented in a dense form, as we do not know beforehand which elements will eventually become non-zero.

We introduce the \emph{adaptive, sparse hyperlayer}, a method for learning a sparse transformation, paramatrized \emph{sparsely}: as index-tuples with associated values. To overcome the lack of gradients from such a discrete structure, we introduce a method of randomly sampling connections, and backpropagating over the randomly wired computation graph.

To show that this approach allows us to train a model to competitive performance on real data, we use it to build two architectures. First, an attention mechanism for visual classification. Second, we implement a method for \emph{differentiable sorting}: specifically, learning to sort unlabeled MNIST digits, given only the correct order.

\end{abstract}

\section{Introduction}

Deep learning is most effective when the transformations we apply to our data are in some way constrained, based on the knowledge we have about our task. For instance, we know that images have a highly local structure, so instead of using fully connected transformations, we use convolutions, connecting input units to hidden units based on locality in the data.

In other words, we replace a densely connected transformation by a sparsely connected one. Sometimes, we know the sparse structure of this transformation beforehand (which connections should exist, and which should not). In other cases, we do not know what structure works best, or worse, the structure should be derived from the data every time the transformation is applied, as in attention mechanisms, or sorting networks. Such sparse transformations are commonly parametrized by dense tensors of equal size, using methods such as sparsemax \cite{martins2016softmax}, or Sinkhorn \cite{mena2018learning} operators to enforce sparsity in the resulting matrix

We propose a different approach: instead of parametrizing a sparse tensor by a dense one, we use a sparse parametrization. 

For sparse tensors with fixed structure, sparse descriptions are common: we store every nonzero element in the tensor as an index-tuple together with its values. For all other elements, we assume value zero. This scheme is well-established, and allows the \emph{values} of the sparse tensor to be learned efficiently through backpropagation. However, if we want to learn the \emph{structure} of the tensor  as well, we run into a problem: since the index tuples are necessarily integers, we cannot derive a meaningful gradient to use in gradient descent. Approaches like the Gumbel-softmax trick \cite{jang2016categorical} or the \texttt{REINFORCE} algorithm \cite{williams1992simple} can be used to approximate such a gradient, but these require a categorical distribution over the space of all integer index-tuples, resulting in another dense parametrization.

Instead, we parametrize each integer index tuple by a distribution over a continuous space containing all integer index tuples. We then sample index tuples randomly, and distribute the corresponding value over these by the associated probability. The gradient propagates back only through these mixed values, and the discrete index tuples are taken as constants. 

We call this mechanism a \emph{sparse layer}. The parameters of the sparse layer can be trained as free parameters, or they can be the output of another network, a \emph{source} network. This turns the sparse layer into a \emph{sparse hyperlayer} (or hypernetwork) \cite{ha2016hypernetworks}. The source network can simply derive the sparse parameters from another parameter space to constrain or bias the learned sparse tensors (for instance to allow only tensors implementing convolutions to be learned). The source network may also derive the sparse parameters from the current input, allowing an \emph{adaptive sparse hyperlayer}, to implement, for instance, an attention mechanism.

We perform several experiments to validate this approach. First, in order to visualize the behavior of the model under gradient descent, we simply learn the identity matrix from batches of normally distributed vectors. While this is a simple problem we show that a comparable \texttt{REINFORCE}-based model cannot solve the task for anything but very small vectors.

Second, to test the model on a real problem, we compare to the Recurrent Attention Model \cite{mnih2014recurrent} on the Cluttered MNIST-100 dataset, a challenging dataset that embeds MNIST digits in $100 \times100$ images with added noise. We show that our approach outperforms the RAM model, while being end-to-end trainable with plain backpropagation.

Finally, we turn to the problem of \emph{differentiable sorting}. We use the principle of the sparse hyperlayer to implement a differentiable version of \texttt{quicksort}. While the sorting algorithm itself contains no trainable parameters, it allows us to sort a set of $n$ tensors by $n$ scalar \emph{keys} in such a way that we get a gradient over the keys which can be backpropagated to the network used to compute the keys. We test this approach by passing MNIST digits through a randomly initialized ConvNet to derive a key, sorting by the keys, and backpropagating to train the ConvNet.

All code is available on-line under, an open-source license.\footnotemark

\footnotetext{\url{https://github.com/MaestroGraph/sparse-hyper}}

\section{The adaptive, sparse hyperlayer}

\begin{figure}[hbt]
\hspace{-0.2\textwidth}
  \includegraphics[width=1.4\textwidth]{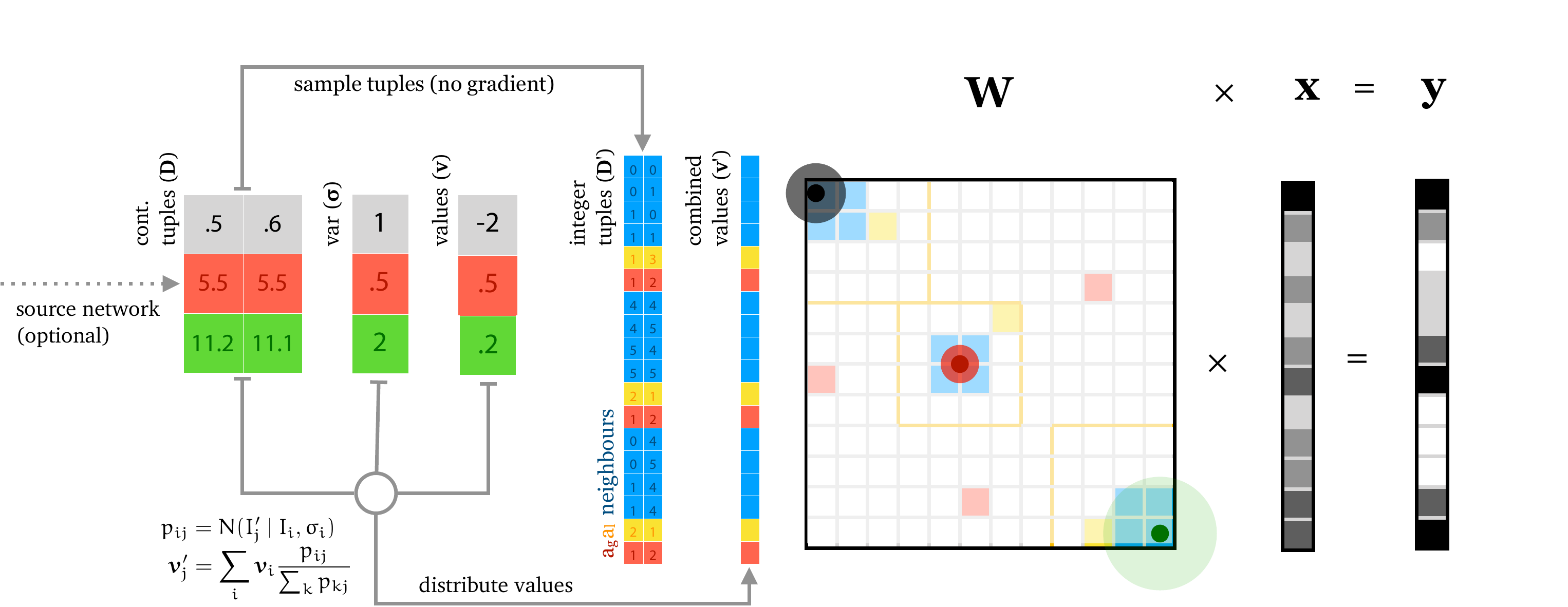}
  \caption{Schematic representation of how the sparse layer is constructed for vector input $\x$ and output $\y$.}
  \label{figure:sl-diagram}
\end{figure}

We define our model as a transformation of one tensor $x$ to another $y$, both of arbitrary rank and shape. The most common transformation of one vector into another by a matrix of weights follows as a special case.

\paragraph{Notation} We use the notation $\mathbold{T}[i, j, k]$ for the element of tensor $\mathbold{T}$ that is indexed by indices $i$, $j$ and $k$. We will use an interpunct to represent subtensors, for instance, $\mathbold{M}[\cdot, j]$ represents the $j$-th column of matrix $\mathbold{M}$. 

Unless otherwise noted, operations between tensors are computed element-wise. Multiplication between matrices of unequal size represents matrix multiplication.

We use $[f..t]$ to indicate a range of integers (including both endpoints) and $\x[f..t]$ to represent a subvector of vector $\x$ indexed by that range (similarly for tensors). For tensor $\X$ and binary tensor $\I$ of the same size, $\X[\I]$ refers to the set of those elements for which $\I$ is 1 and $\X[\neg \I]$ refers to the set of elements for which $\I$ is 0. If this set can be arranged into a tensor in a straightforward way (for instance if $\x$ is a vector), we will take $\X[\I]$ to represent that tensor.

\subsection{Model}
\label{section:model}

Let $\x$, the \emph{input}, be a tensor in $\R^{i_1 \times \ldots \times i_n}$ and let $\y$, the \emph{output}, be a tensor in $\R^{o_1 \times \ldots \times o_m}$. A \emph{sparse layer} is a compressed representation of a weight tensor $\W$ representing a fully connected layer, so that the layer computes the operation $\W \cdot \x = \y$, where the dot represents a tensor contraction defined below, which equals matrix multiplication if $\x$ and $\y$ are vectors. 

The shape of $\W$ is determined by the shapes of $\x$ and $\y$ as $\W \in \R^{h_1 \times \ldots \times h_r}$ with $r = n + m$. The specific tensor contraction computed is defined as follows: let $a_1, \ldots, a_n$ represent indices of $\x$, and let $b_1, \ldots, b_m$ represent indices of $\y$. We then define $\W \cdot \x = \y$ as:
\[
\y[b_1, \ldots, b_m] = \sum_{a_1, \ldots, a_n} \W[b_1, \ldots, b_m, a_1, \ldots, a_n] \cdot \x[a_1, \ldots, a_n] \label{eq:tensormult}
\]

That is, for each element of $\y$ indexed by $b_1, \ldots, b_m$, $\W$ contains a subtensor $\W[\cdot, \ldots,\cdot, b_1, \ldots, b_m]$ of the same shape as $\x$, and $\y[b_1, \ldots, b_m]$ is the dot product of that subtensor and $\x$. In network terms, $\W$ represents a fully connected layer: for every input node there is one weight in $\W$ for every possible connection to an output node. 

We assume that the rank of the tensor and the output dimensions stay constant over all inputs. The input dimensions can vary if eager execution is used, but not the input rank.

We will assume that $\W$ is sparse. Our aim is to learn both its structure and values using a sparse parametrization: that is, we enumerate the index tuples of its nonzero elements and assume that anything not enumerated has value zero. To do this in a differentiable way, we define a matrix $\D \in \R^{k \times r}$ as real-valued representatives of the index tuples, and two length-$k$ vectors $\sig$ and $\bv$. Broadly, the rows of $\D$ encode points near likely integer index tuples. $\sig$ encodes the corresponding certainty: the smaller $\sig[i]$, the more certain we are that $D[i, \cdot]$ is in the right place. Finally, $\bv_i$ is the value of the element in the matrix at index $\D[i, \cdot]$.

We map $\D$ to the space of legal index tuples, by applying a sigmoid function and scaling to the hyper-rectangle containing the tuples. Each scalar $\sig[k]$ is passed through a softplus function to make it positive, expanded to a length-$r$ vector, and multiplied by the dimensions of $\W$, resulting in a vector $\sig_k$, so that $\text{diag}(\sig_k)$ represents a convariance matrix that covers an equal proportion of available index tuples in each direction. \footnotemark 

\footnotetext{We also add a small value $\tau$ (a hyperparameter), set to 0.1 in all experiments, to stop the variance from converging too close to zero (which occasionally helps to avoid local minima).}

To form a sparse matrix based on these parameters, we generate several \emph{integer} index tuples from each continuous index tuple $\D[i, \cdot]$:
\begin{itemize}
\item The $2^r$ integer tuples nearest to $D[i, \cdot]$ are always included.
\item We sample $a_l$ \emph{local} index tuples uniformly from a region of size $l_1, \ldots, l_r$ around $\text{round}(d_j)$. If the region overlaps the bounds of the allowed index tuples it is aligned to the bound.
\item We sample $a_g$ \emph{global} index tuples uniformly over all possible integer index tuples.
\end{itemize}
These three sets of tuples are concatenated into an integer valued matrix $\D' \in \R^{k(2^d+a_l+a_g)\times r}$. 

Each sampled integer index tuple $d_j = \D'[j, \cdot]$ receives a  \emph{proportion} from each continuous index-tuple $d_i =\D[i, \cdot]$ defined as $p_{ij} = N(d_j \mid d_i, \sig_i)$. If any index tuple occurs more than once, the proportion of the duplicates is set to zero (algorithm given in the appendix). The proportions are then normalized so that the contribution of each continuous index tuple sums to one over all integer index tuples:
\[
p'_{ij} = p_{ij} / \sum_{k} p_{ik}
\]

We then compute a value $\bv'[j] = \sum_i p'_{ij}\bv[i]$ for each sampled integer tuple. Effectively, each generated index tuple is assigned a part of $\bv[i]$ proportional to its probability under $N(d_i, \sig_i)$. 

Together, the generated integer index tuples $\D'$ and their values $\bv'$ define a sparse matrix $\W$. The precise procedure is given in Algorithm~\ref{algorithm:sample-w}.

\begin{pseudo}
\caption{Sample a sparse matrix $W$ with integer index tuples from sparse parameters $\D$ and $\bv$.}
\label{algorithm:sample-w}

\textbf{given}:\\
\tab $\D \in \R^{k \times r}; \sig, \bv \in \R^k$\\
\tab scalars $a_g, a_l, (l_1, \ldots, l_r)$\\

\textbf{initialize} $\D' \in \N^{k (2^r + a_l + a_g) \times r}$, $\bv' \in \R^{k(2^r + a_l + a_g)}$ \\ 
\\
\textbf{for} $i \in [0 .. k]$: \\
\tab $\bd_i \leftarrow \text{sigmoid}(\x) \cdot[h_1, \ldots, h_r]$ \\
\tab $\sig_i \leftarrow$ softplus$(\sig[i] + 2) \cdot[h_1, \ldots, h_r] \cdot 0.1 + \tau$\\
 
\tab $\D'[i..i+2^r, \cdot] \leftarrow$ the $2^r$ integer index tuples nearest to $\bd_i$\\
\tab $\D'[i+2^r..i+2^r+a_l, \cdot] \leftarrow a_l$ tuples uniformly over $(l_1, \ldots, l_r)$ around $\bd_i$\\
\tab $\D'[i+2^r+a_l..i+2^r+a_l+a_g, \cdot] \leftarrow a_g$ tuples uniformly over all tuples \\

\tab \textbf{for} $i \in [1..k]$, $j \in [1..2^r+a_g+a_l]$:\\
\tab\tab\tab $p_{ij} = N(\D[j, \cdot] \mid \bd_i, \sig_i)$\\
\\
\tab \textbf{for} $j \in [1..2^r+a_g+a_l]$:\\
\tab\tab $\bv'[j] = \sum_i \bv[i] \frac{p_{ij}}{\sum_k p_{ik}}$\\
\\
\textbf{return} $\D'$, $\bv'$\\
\end{pseudo}

We can now perform our transformation as
\[
\y = \W \cdot \x 
\]
using standard sparse-times-dense matrix multiplication. During backpropagation, we compute the gradient \emph{only} over $\bv'$, taking $\D'$ as a constant. This still allows us to learn the values of $\D$ and $\sig$, since $\bv'$ is computed using all parameters.
%
%We will apply the sparse layer in two forms. First, treating $D$, $\sig$ and $\bv$ simply as parameters and learning a \emph{non-adaptive sparse (NAS) layer}. 
%
%Second, we will use a  network to produce $D$, $\sig$ and $\bv$, usually from the model's input (this may be $x$ or it may be some data from which $x$ is produced). This gives us an \emph{adaptive sparse hyperlayer (or ASH-layer)} which can tune the sparsity of its structure. 
%
%Such a hypernetwork is not only useful to produce adaptivity. Even when we only want to learn a static transformation, a hypernetwork can be used to constrain the structure of $\W$. Examples include learning triangular matrices, block matrices, or rotation matrices.

For a matrix-by-matrix multiplication, where $\W$ , $\x$ and $\y$ are all matrices, the algorithm can be applied without modification, although in that case it does not generalize in a simple manner to higher-dimensional tensors.

In some cases both the distribution on the index tuples and the method of sampling index tuples can be tailored to the task at hand. Section~\ref{section:sorting}, provides an example.
 
\section{Experiments}

To show that the sparse layer can learn sparse transformations effectively, and can be employed to solve practical and challenging problems, we test it in three experiments.

\begin{figure}[tbh]
\hspace{-0.2\textwidth}
  \includegraphics[width=1.4\textwidth]{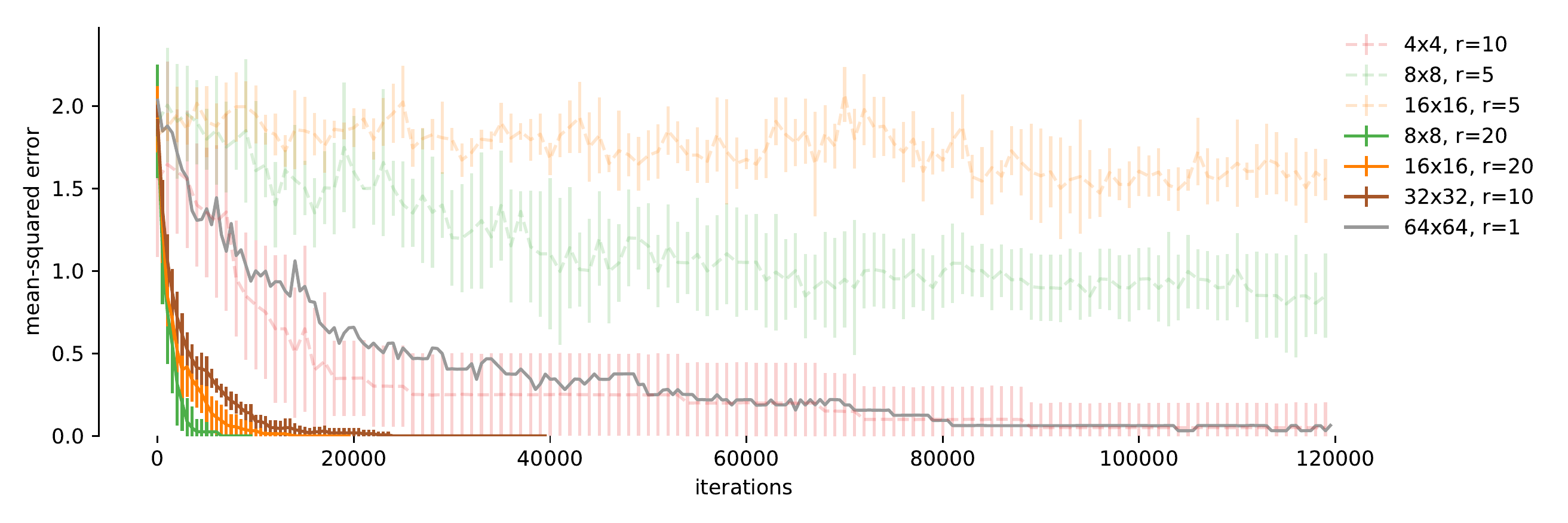}\\  \caption{Results of the identity experiment, comparing the performance of the \texttt{reinforce}-based baseline (dashed) to that of the sparse layer (line). Color indicates size. Error bars denote the \emph{standard deviation} over $r$ runs. The first three sparse layer experiments were stoped after 10\,000, 20\,000 and 40\,000 iterations respectively.}
  \label{figure:identity-results}
\end{figure}

\subsection{Learning the identity matrix}

As a simple toy example, to illustrate the way the sparse layer behaves, we learn the identity matrix: the input is a size-$n$ vector $\x$ from $N(0, \mathbold{I}^n)$, we compute the mean-squared-error loss between $\x$ and $\W x$ with $\W$ a sparse layer (with no source network), and backpropagate as show above. 

We also train a \texttt{REINFORCE}-based model as a baseline, which parametrizes $\W$ in the same way. Each epoch, we sample from each probability distribution representing an index tuple, a single continuous value. We round this value to the nearest integer index tuple, and construct a sparse matrix $\W$ from the resulting integer index tuples. We use \texttt{REINFORCE} to backpropagate through the sampling step.

We aim to sample a log-linear number of index-tuples in $n$ (versus the quadratic total). To this end we set $a_l$ and $a_g$ so that $a_l + a_g < 2\log_2 n$.~ \footnotemark

\footnotetext{With only four sizes tested, we should be reluctant to conclude that larger identity matrices can also be trained to convergence with a sparse number of samples. We leave this question to future research.}

We evaluate by rounding the continuous index tuples to the nearest integer values and computing the loss over 10\;000 newly sampled instances. Each point in the graph represents such an evaluation.

\begin{figure}[tbh]
\centering

  \includegraphics[width=0.35\textwidth]{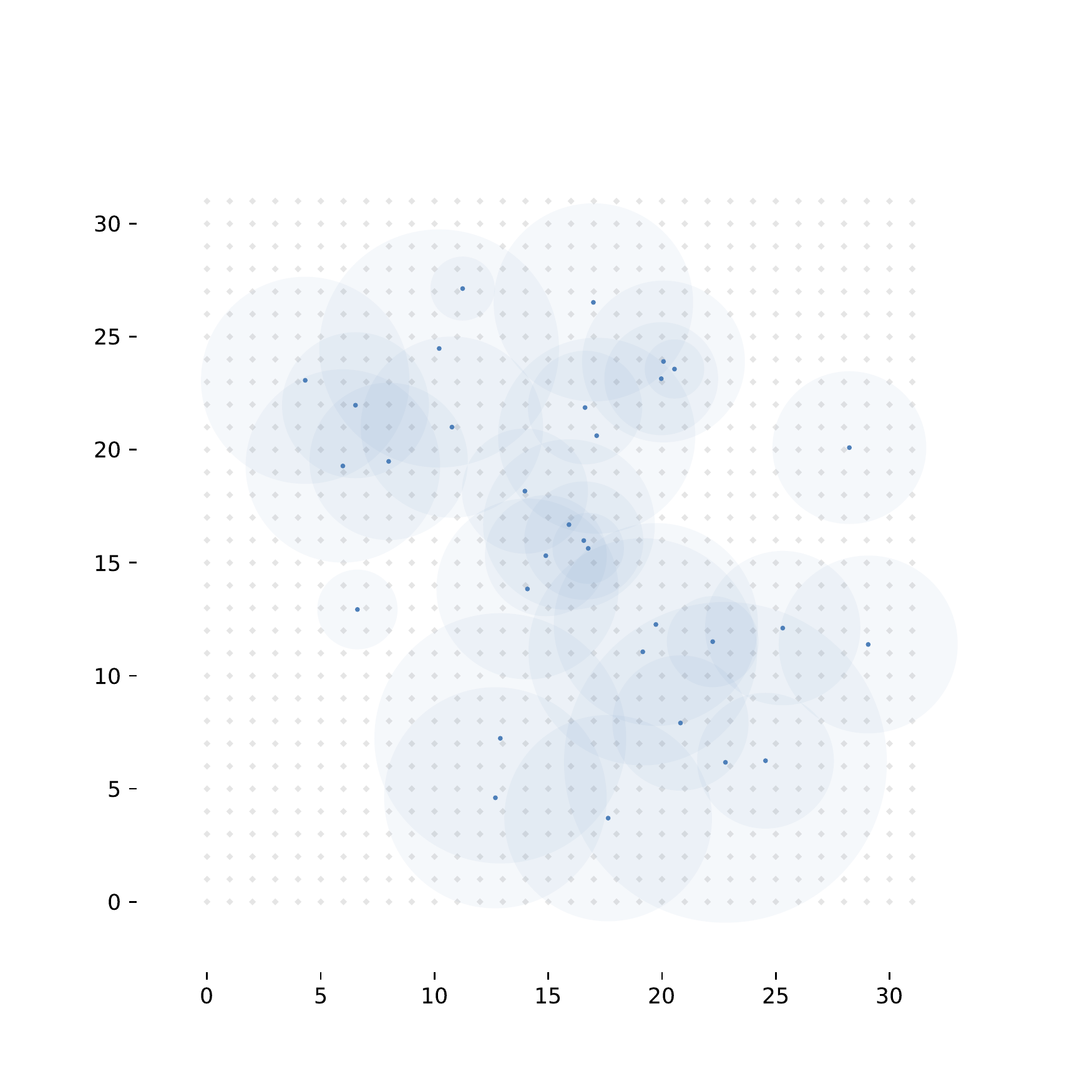}\hspace{-0.8cm}
  \includegraphics[width=0.35\textwidth]{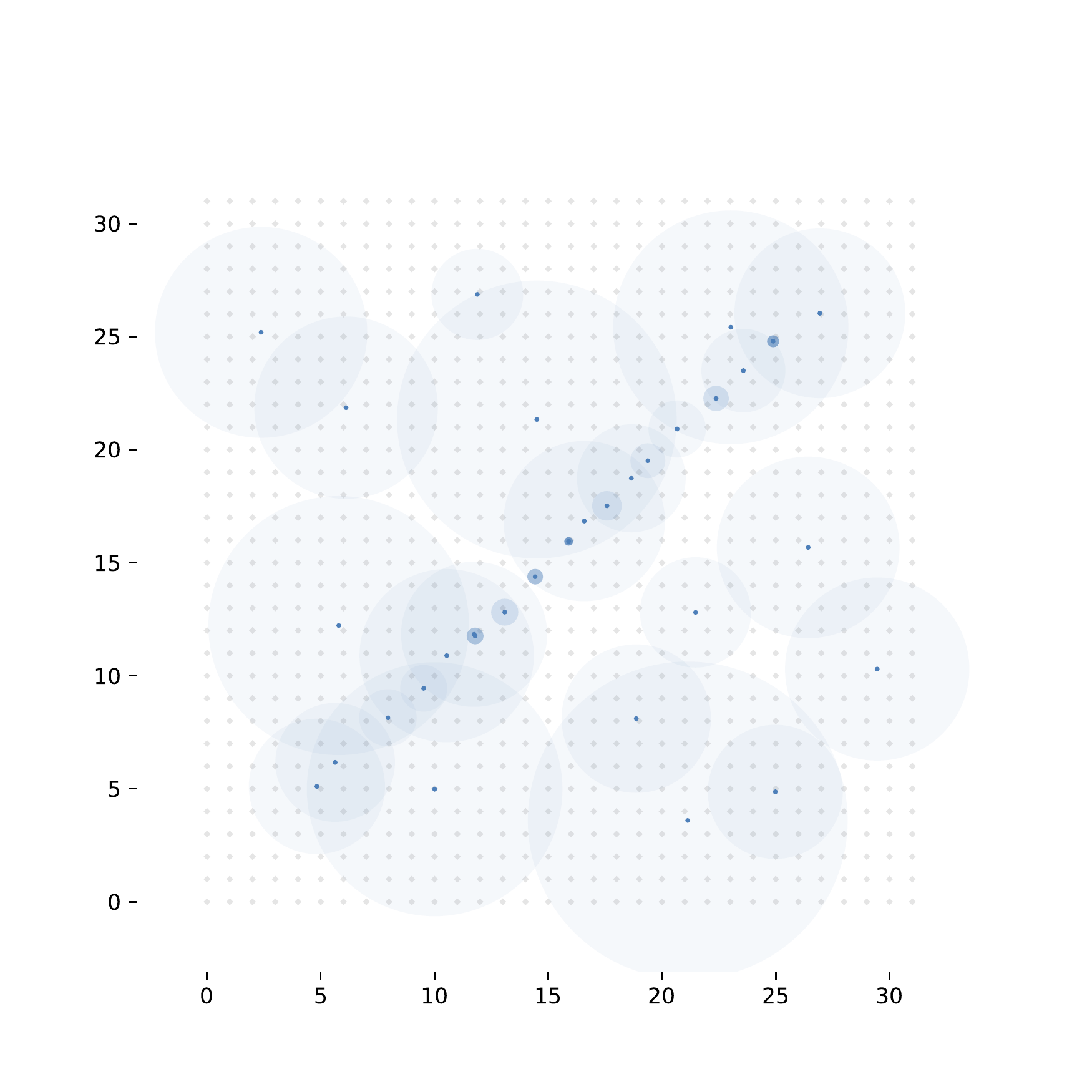}\hspace{-0.8cm}
  \includegraphics[width=0.35\textwidth]{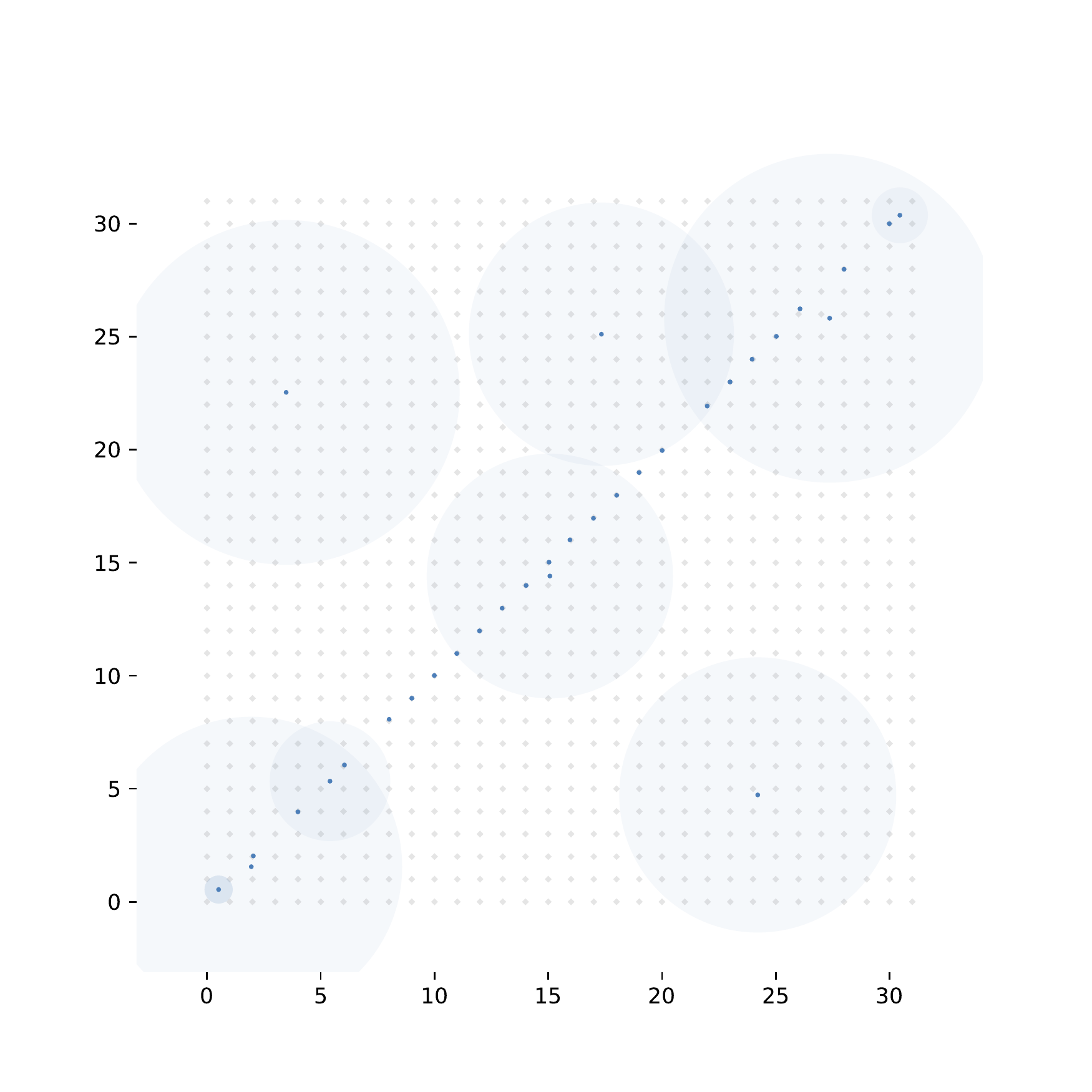}\hspace{-0.8cm}
  \includegraphics[width=0.52\textwidth]{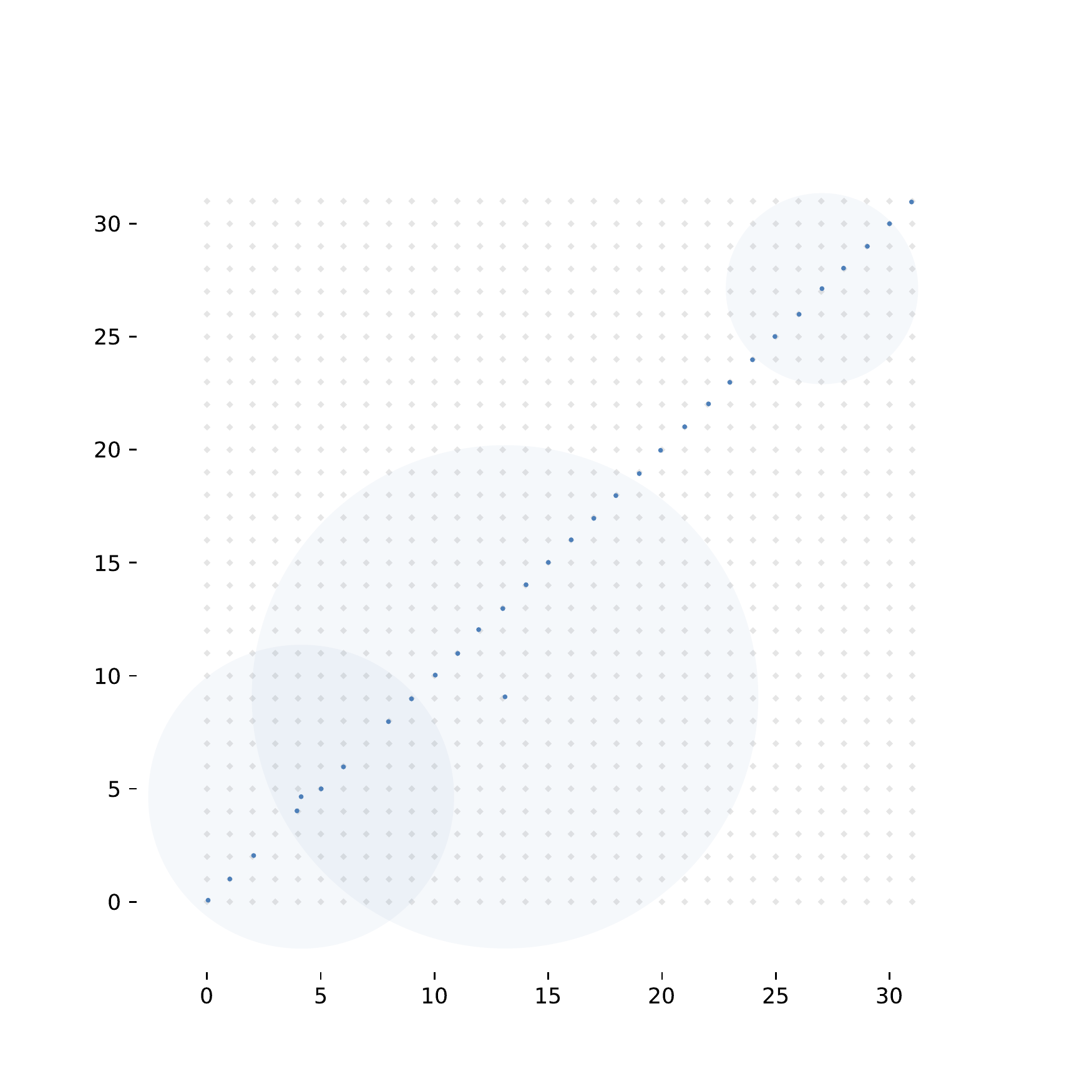}\hspace{-0.8cm}
  \includegraphics[width=0.52\textwidth]{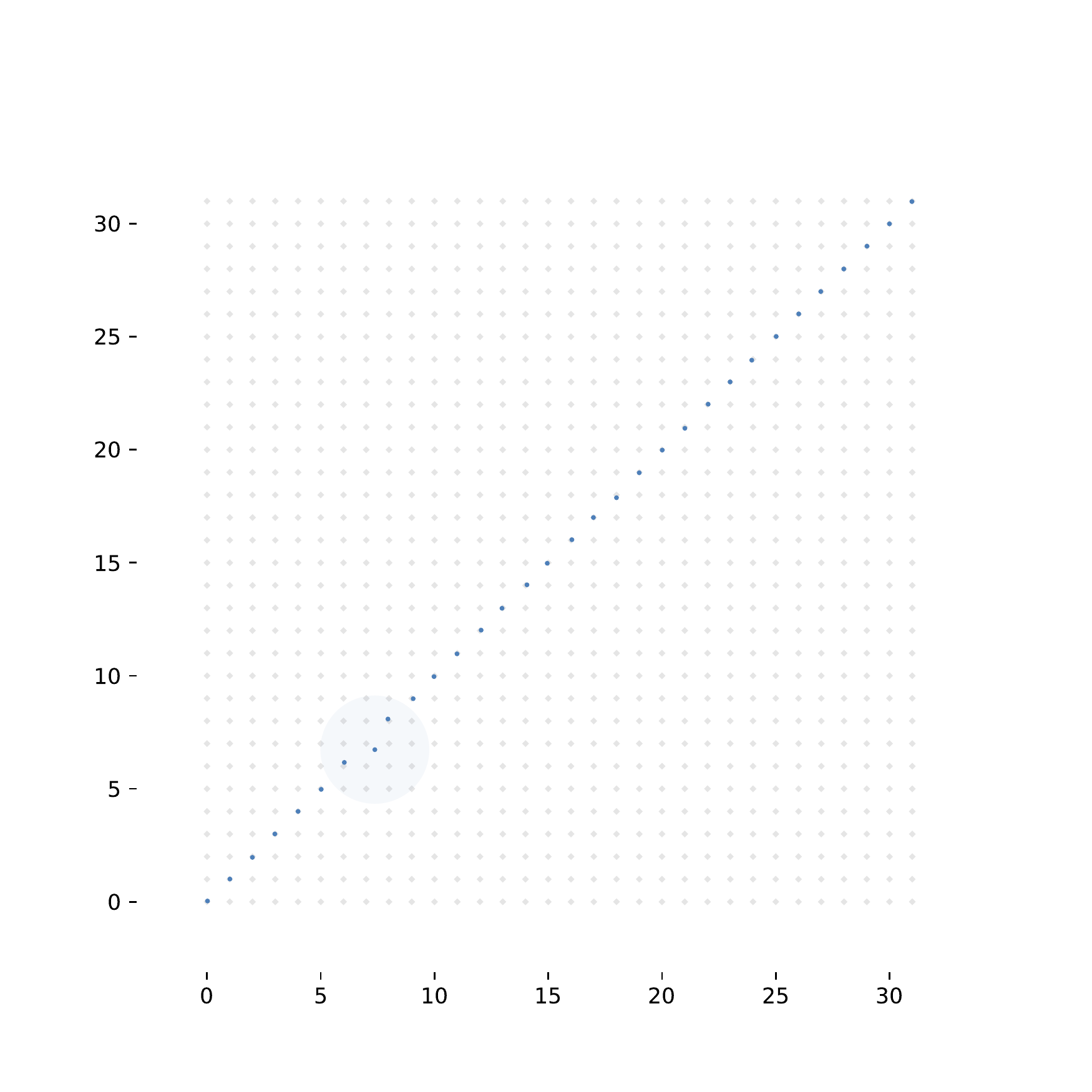} 

\caption{Visualization of the sparse layer after 0, 1\,000, 5\,000, 9\,000 and 13\,000 batches, in the size-32 experiment. Solid blue circles indicate the means of the continuous tuples. Transparent circles represent the corresponding variance. Gray diamonds indicate the locations of integer index tuples.}
  \label{figure:identity-viz}
\end{figure}

Figure~\ref{figure:identity-results} shows the results. We see that the baseline seems to work, if slowly, up to a size of 8, but fails for size 16. The sparse layer is able to train to convergence up to size 64. For greater sizes, we still see the model move towards the correct solution, but training to full convergence requires more resources than were available. As in the size-32 example shown in Figure~\ref{figure:identity-viz}, most index tuples move to the diagonal quickly, but it takes more iterations for the final ones to find their place.

A more elegant use of \texttt{REINFORCE} may lead to better results, but this experiment does highlight the fundamental difference between the two approaches. Consider for instance, that for $n=4$, \texttt{REINFORCE} must take 4 samples, one for each index tuple. The downstream loss is then used to approximate the gradient for all four of these `actions', with each stochastic node receiving the same loss. The loss is distributed by the individual \emph{probability} of each sample but not by the individual \emph{quality} of each sample.

Conversely, the sparse layer, after sampling, functions as a single computation graph: those samples that hit the right index tuples receive positive feedback and those that hit the wrong ones receive negative feedback. The backpropagation algorithm can distribute the loss accurately back down the computation graph as it was used in the forward pass.

\subsection{Attention}

\begin{table}[hbt]
\centering
  \begin{tabular}{l|cc}
   model & error \\
    \hline
   RAM (best reported) & 8.11 \% \\
   ours (1 glimpse, $32 \times 32$) & 0.85 \% \\
   ours (8 glimpses, $12 \times 12$) & 1.01 \% \\
  \end{tabular}
  \caption{The results of the attention experiment. The baseline consists of the same ConvNet we use as a source network extended with a classification layer (details in the appendix).}
  \label{table:attention-results}
\end{table}

As the previous experiment shows, a large, unconstrained sparse layer, can take a long time to converge. The model becomes most effective when we use a source network to constrain which sparse tensors the network is able to represent. We start with an extreme example in the form of \emph{attention}. We are given a large input image with a smaller region of interest which determines its class. The task of our layer is to extract the region of interest, and pass it to a classification network. 

In this setting, our input $\x$ is an image tensor containing 3 dimensions: channels $c=1$, height $h_\x=100$ and width $w_\x=100$. Our sparse layer transforms this to a smaller $k \times k$ image $\y$ (which we'll call a \emph{glimpse}, after \cite{mnih2014recurrent}) also represented as a 3-tensor. The sparse layer $\W$ for this transformation is therefore a 6-tensor with dimensions $(c, k, k, c, h_\x, w_\x)$.

It may be tempting to have the source network simply learn the paramaters of $\W$ without constraint. However, doing this results ``feedforward behavior''. That is, it simply selects a black pixel to transmit a 1 and a white pixel to transmit a 0. For the network to behave as an an attention mechanism, we need to constrain it much further. 

First, we learn only along two of the dimensions of $\W$: $h_\x$, $h_\y$. For the others we fix their values so that each element is determined by one element in $\W$. In network terms: each node in the output receives a single incoming connection, and we learn only which node in the input it should be connected to.\footnotemark~The values $\bv$ are learnable, but shared between all connections in a glimpse.

\begin{figure}[bth]
  \centering
  \includegraphics[width=\textwidth]{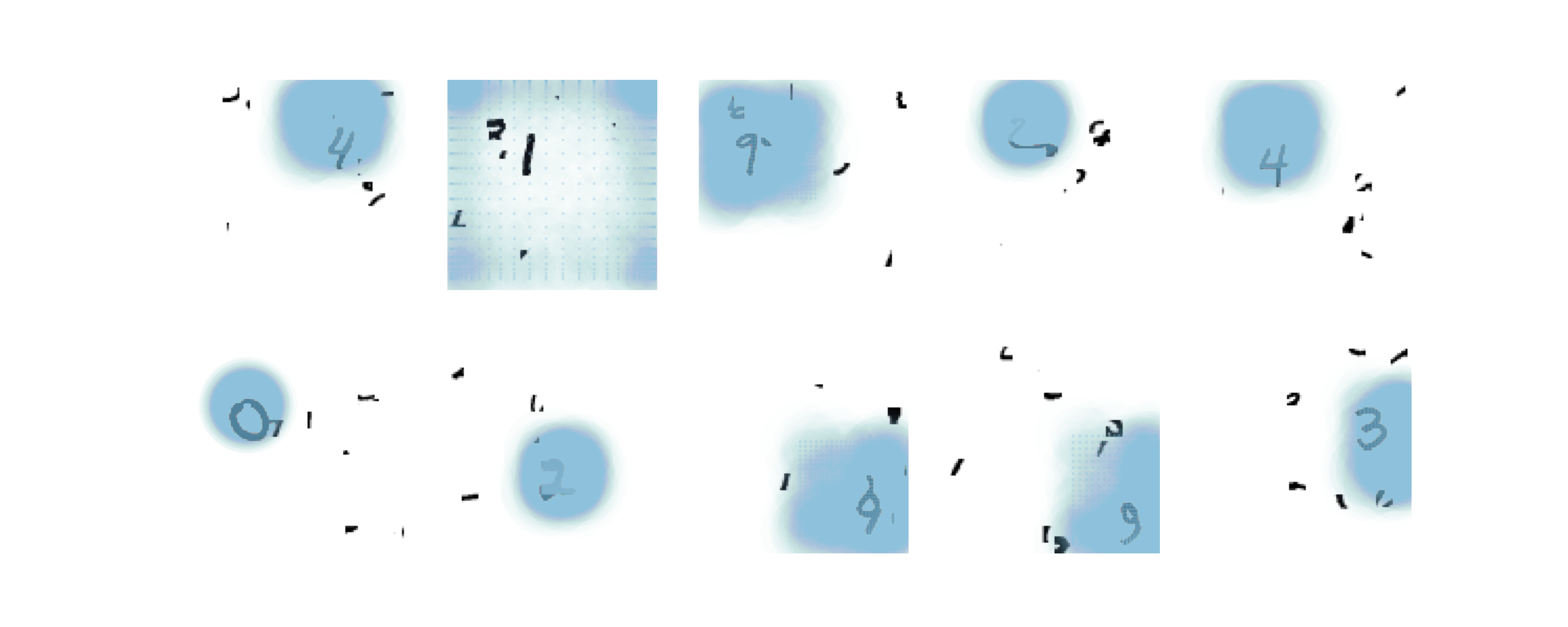}
  \includegraphics[width=\textwidth]{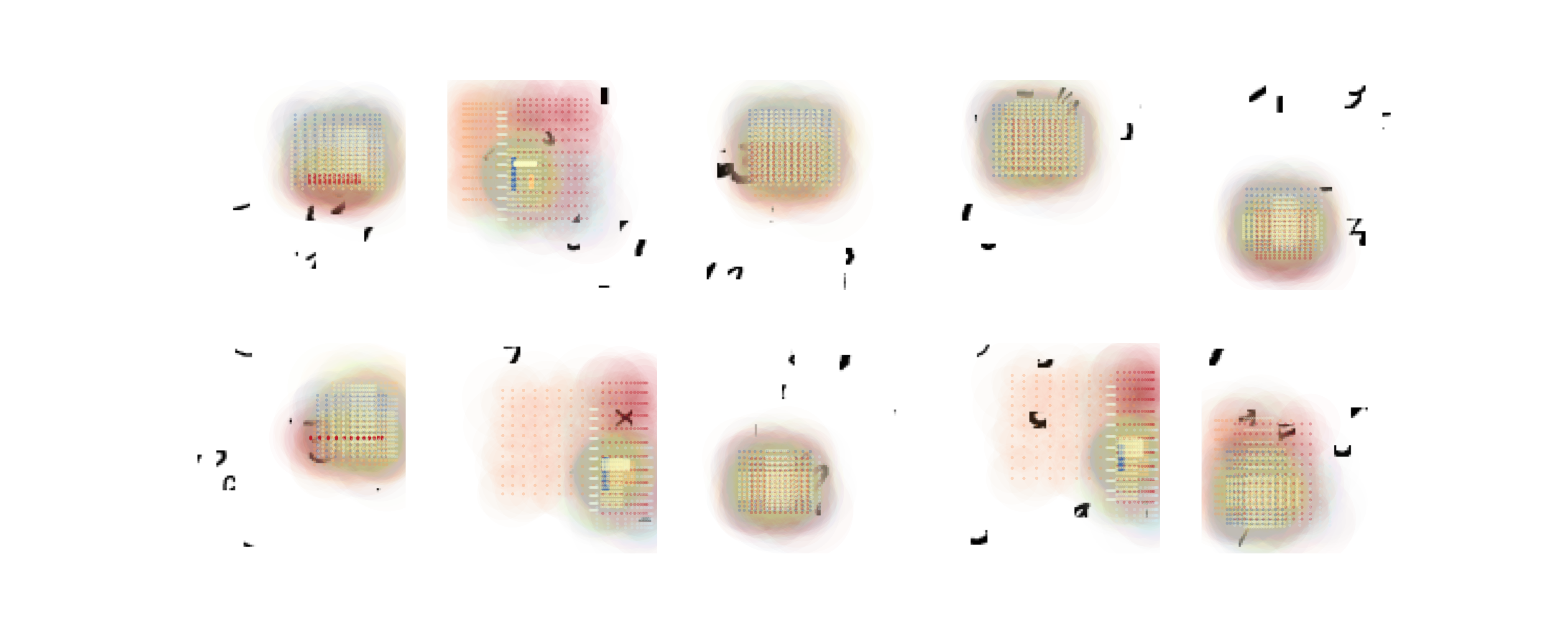}
  \vspace{-0.5cm}
  \caption{The glimpses after the final epoch, for the 1-glimpse network (above) and the 8-glimpse network (below). Positive values are coded, blue, negative values are coded red. Values near zero are coded yellow.}
  \label{figure:glimpses}	
\end{figure}

For convenience of implementation, we implement each glimpse in a separate sparse layer. Though it may be beneficial to combine the glimpses in a single tensor, we have not yet tested this approach.

\footnotetext{For multiple channels, each input channel is connected only to a corresponding output channel, but we use only single-channel images in this experiment.}

A source network (architecture given in the appendix) reads the input image, and produces, for each glimpse, 4 values, representing the lower left corner of the bounding box for the glimpse and the upper right. We then compute $k \times k$ equally spaced points in this bounding box and connect these points in the input to their corresponding $k \times k$  points in the output. The sigmoid and rescaling are applied afterwards.

We then proceed as described in Section~\ref{section:model}, sampling integer index tuples for each continuous tuple in each glimpse. We feed the resulting glimpses to a 2-layer MLP network which maps the glimpses to a class.

We test this approach with the $100 \times 100$ MNIST-cluttered dataset from\cite{mnih2014recurrent}. We resample the data,\footnotemark creating a training set of 60\,000 images and a test set of 10\,000 images. All hyperparameters were chosen using a validation split, and the test set was only used once.

\begin{figure}[tbh]
\hspace{-0.2\textwidth}
  \includegraphics[width=1.4\textwidth]{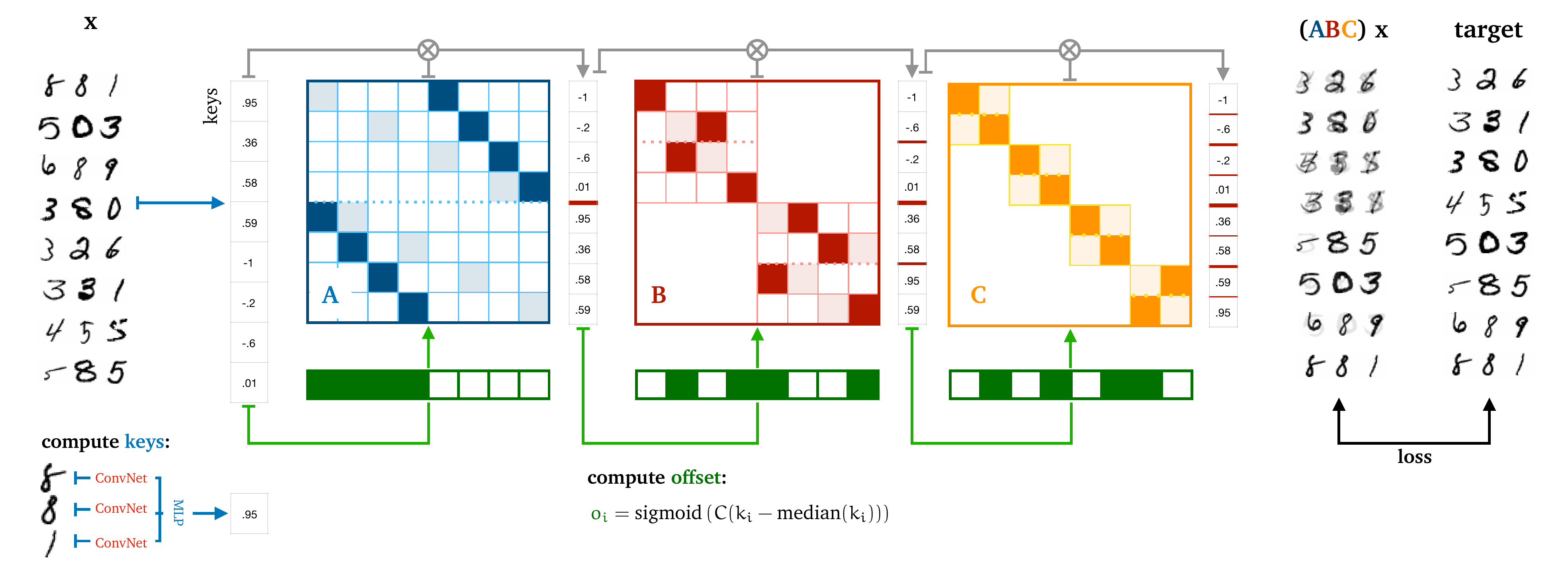}
  \caption{Schematic representation of differentiable \texttt{quicksort}. A \emph{key network} maps each 3-digit number to a scalar key (applying the same ConvNet to each digit). The keys are used to create three half-permutation, which compose to form a full permutation required to sort the keys. We apply this permutation to the data $\x$, and backpropagate the loss to train the network that generated the keys.}
  \label{figure:sorting}
\end{figure}

We test two configurations: a network with a single glimpse of $32 \times 32$, and a network with 8 glimpses of $12 \times 12$ (following \cite{mnih2014recurrent}).

\footnotetext{Using the script provided at \url{https://github.com/deepmind/mnist-cluttered}. The sampled data can be downloaded from \url{https://github.com/MaestroGraph/sparse-hyper/}}

Table~\ref{table:attention-results} shows the results. Figure~\ref{figure:glimpses} visualizes the learned glimpses. We note, that while the 1-glimpse network classifies with high accuracy, it does not always behave exactly as expected from an attention mechanism. In many cases it selects only part of the digit. We suspect that this is a kind of intermediate between a feedforward network and an attention mechanism: by selecting a specific part of each digit, the source network communicates upstream part of the classification information, lightening the load of the upstream network. Further research is required to study this behavior.
The 8-glimpse network seems to behave better, focusing more on the entire digit, although this does not lead to better classification accuracy.

Without further research, we will not claim that this is a competitive attention mechanism, but this experiment does show that the adaptive, sparse hyperlayer can be trained effectively through backpropogation, in a realistic setting.

\subsection{Differentiable sorting}
\label{section:sorting}

Finally, we turn our attention to the problem of \emph{differentiable sorting}, or equivalenty, of \emph{learning permutations}. Some early successes in this area have come in the form of pointer networks \cite{vinyals2015pointer} and Gumbel-sinkhorn networks \cite{mena2018learning}.

These approaches both, to some extent, attempt to bias a parametrized network towards learning sorting behavior. Our approach is more modular. We first define a \emph{key network} over the instances to be sorted. The key network assigns a single, real value to each instance. We then implement, using adaptive sparse layers, a \emph{differentiable} version of \texttt{quicksort} \cite{hoare1962quicksort} that allows us to compare the output to the correctly sorted version, and backpropagate the loss to train the key network. We call this approach more modular, because the network that performs the sorting contains no parameters. The permutation is entirely parametrized by the key network, and is easily interpretable (the order of the keys determines the learned permutation). For a schematic illustration of the following, see Figure~\ref{figure:sorting}.

In order to implement \texttt{quicksort} using sparse layers, we first introduce the concept of a \emph{half-permutation}. A \emph{first-order} half permutation is a permutation which arranges the given elements into the top or bottom half of the output, but preserves their ordering otherwise. More precisely, let $\bo$ be a length-$n$ binary vector with equal numbers of $0$ and $1$ elements, and let $\x$ be an input vector. The 0-order half permutation defined by $\mathbold{o}$ re-arranges its input so that the result is the concatenation of $\x[\bo ]$ and $\x[\neg \bo ]$. We denote the permutation matrix implementing this half-permutation $P^1(\bo)$.

Note that if we define 
\[
\bo[i] = \begin{cases}
	0 & \text{if } \x[i] < \text{median}(\x) \\
	1 & \text{otherwise}	
 \end{cases}
\]
then $P^1(\bo)$ implements the first step of \texttt{quicksort}: it divides the input in to two ``buckets'' (the top and bottom half of the output), otherwise maintaining its order. 

For subsequent steps we define higher order half-permutations. A $d$-th order half-permutation is a permutation of $n$ elements, where $n$ is divisible by $2^d$, defined by permutation matrix $P^d(\bo)$: a block-diagonal matrix, with $(d-1)$-order half-permutions along its diagonal. Specifically, the matrix covering elements $\left[m..m+n2^{-(d-1)}\right]$ of the diagonal, is $P^{d-1}\left(\bo\left[m..m+n2^{-(d-1)}\right]\right)$.

We can define $\bo$ for the $d$-th step of \texttt{quicksort} as follows. Divide the range $[1..n]$ into chunks of size $n2^{-(d-1)}$ and let $r(i)$ be the chunk containing index $i$. We then define:
\[
\bo(x)[i] = \begin{cases}
	0 & \text{if } \x[i] < \text{median}(\x[r(i)]) \\
	1 & \text{otherwise.}
 \end{cases}
\]

Assuming that our input has a length $n$ that is a power of 2, we can now implement \texttt{quicksort}\footnotemark\,
 by composing a total of $\log_2n$ half-permutations. Call the input $\x_0$ and let \begin{align}
\x_i &= P^{i}(\bo(\x_{i-1})) \cdot \x_{i-1} \p \label{equation:sorting} 
\end{align}

\footnotetext{Specifically, \texttt{quicksort}, with median pivots. In normal sorting this is not a popular approach, but in our case, it is crucial to guarantee that a depth of $\log_2 n$ is sufficient to to sort the input.}

After applying $\log_2 n$ half-permutations, $\x_{1+\log_2n}$ contains the sorted input. 

To allow a meaningful gradient to propagate through through this computation, we implement the half-permutations as sparse layers, parametrized by $\bo$. We replace the discrete, integer-based definition of $\bo$ by a continuous one:
\[
\bo' = \text{sigmoid}\left ((\x_i - \text{median}(\x[r(i)]))C \right)
\label{eq:sigmoid}
\]
Where $C$ is a hyperparameter set to 10 in all experiments. Note that as the distance to the median grows (for instance if we learn to place the keys further apart), $\bo'$ converges to $\bo$.

We then use $\bo'$ to parametrize a sparse layer that approximates $P^d(\bo')$ as follows. We sample $a$ random binary vectors ${\bo_i}$ representing valid half-permutations of order $d$ by starting with the vector that represents the identity permutation (note that this is different for each $d$) and permuting the values in each chunk of size $n2^{-(d-1 )}$ randomly.\footnotemark~We also add $\text{round}(\bo')$ to the sample.

\footnotetext{Since permutations are costly to compute in most deep learning frameworks, we sample a matrix of 500\,000 random permutations at initialization, and sample random rows from that.}

We convert each $\bo$ to a sequence of $\br(\bo)$ row indices, representing the nonzero elements of $P^d(\bo)$ (assuming that the columns are indexed sequentially, algorithm in the appendix). Each such sequence $\br(\bo)$ is assigned proportion
\[
p(\bo) = \prod_i \begin{cases} \bo'[i] &\text{if }\bo[i] = 1 \\
1 - \bo'[i] &\text{otherwise.}
 \end{cases} 
\]

Duplicate sequences $\br(\bo)$ are detected by the method described in the appendix, and their proportion is set to zero. We assign this proportion over all integer tuples generated from $\bo$. We combine the resulting index tuples for all binary vectors in the sample into a single sparse matrix, with the assigned proportions as their values.

Finally, we normalize the proportions so that all columns in the resulting sparse matrix sum to 1.

To see how the gradient propagates backwards, it is instructive to imagine the case $n=1$. In this case the key network produces two keys, and the half-permutation is either the  identity matrix $\I$ or $1-\I$. Let's say we get $\I$ (the keys suggest the instances are already ordered) and by distribution over sampled $\bo_i$, the off-diagonal elements become slightly higher than zero and the diagonal elements slightly lower than 1. If these keys are incorrect (the instances should instead be swapped), gradient descent will increase the off-diagonal elements and decrease the diagonal elements. This means decreasing the proportion assigned to $\bo_0 = (0, 1)$ and increasing the proportion assigned to $\bo_1 = (1, 0)$. Since this is the result of Equation~\ref{eq:sigmoid}, gradient descent pushes both keys closer to their median, and towards the center of the sigmoid curve. Eventually they will swap place, and be pushed away from each other.

To test this algorithm we devise the following experiment. We sample $n$ random 3-digit numbers (with $n$ a power of 2) and represent each by three randomly sampled MNIST digits of the correct label. We pass these through a key network which processes each digit individually using the same ConvNet, concatenates their outputs and passes it through an MLP to produce a single scalar key (details in the appendix).

We sort the digits using the scheme above and compare the outputs to the sorted target, using binary cross-entropy as a loss function.\footnotemark~To evaluate, we ignore the sorting network, generating only the key, and compare the permutation required to sort the keys to the permutation required to sort the digits. We report the error (out of 10\,000 samples): the proportion of instances for which the two permutations are not exactly the same. 

Hyperparameters were chosen using a validation split of the MNIST training data, with the test used used only for the final experiment.

\footnotetext{We clamp the output to $(0,1)$ to avoid small floating-point inaccuracies.}

\begin{figure}[tbh]
\hspace{-0.2\textwidth}
  \includegraphics[width=1.4\textwidth]{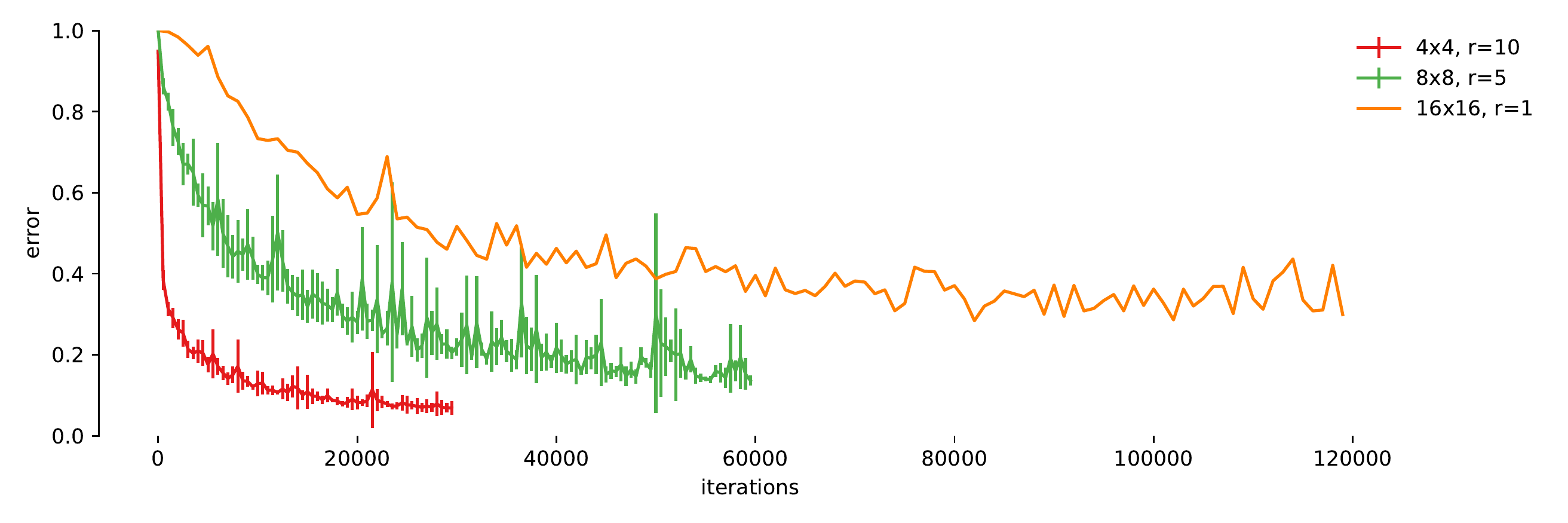}\\  \caption{Results of the sorting experiment. Error bars indicate the \emph{standard deviation} over $r$ repeats of the experiment.}
  \label{figure:sorting-results}
\end{figure}

\paragraph{Intermediate loss} Simply comparing the input to its sorted counterpart works well for short sequences, but for longer sequences we can enrich the loss by comparing intermediate steps. et $d = 1 + \log_2 n$ be the index of the last half-permutation, let $\x_i$ be defined as in (\ref{equation:sorting}), and let $\bt_d$ be the sorted target. Feeding $\bt_d$ through the transposes of the half-permutations in reverse order should retrieve $\x$ (if the keys are in the correct order). Let
\begin{align*}
\bt_{i-1} &= {P^{i}(\bo(\x_{i-1}))}^T \cdot \bt_{i} \p
\end{align*}
We now define the loss as the binary cross-entropy between $\x_i$ and $\bt_i$ summed over all $i \in [1..d]$.

\begin{figure}[tb]
  \hspace{-0.23\textwidth}
  \includegraphics[width=1.4\textwidth]{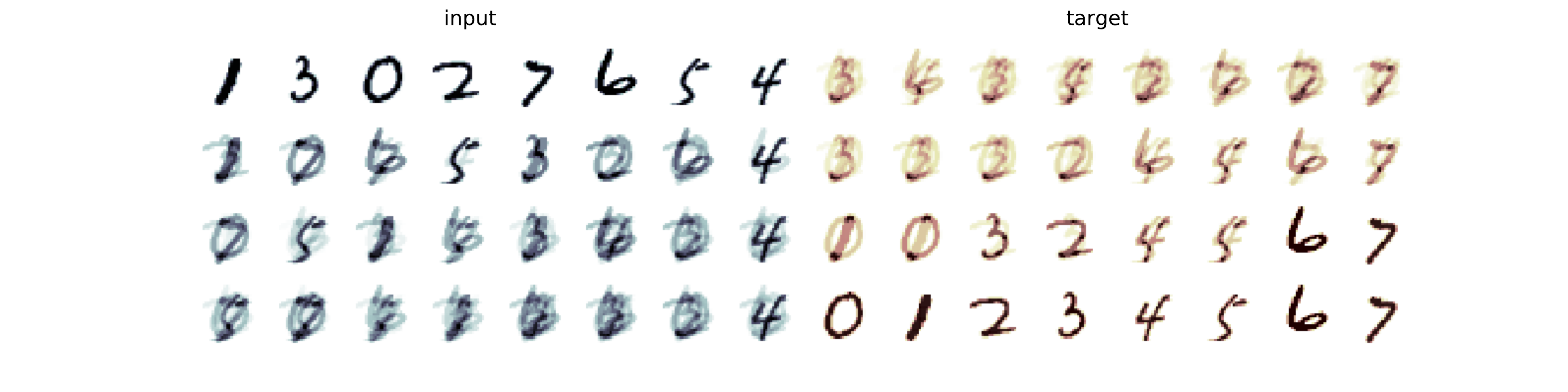}
  \vspace{-1cm}
  \caption{An illustration of the intermediate loss for the task of sorting a sequence of 8 single-digit numbers. The left side shows the input as it is transformed (in the first step of training). The right side shows the target being reverse-sorted.}
  \label{figure:intermediate}
\end{figure}

Figure~\ref{figure:intermediate} illustrates the intermediate steps at the start of learning. Note how the intermediate losses provide guidance at each step: the final target row shows that the first element should be 0, and the preceding target row shows that the first two elements should each be mixtures of a 0 and a 1. The row above shows that the first four elements (the first ``bucket'' in the \texttt{quicksort} algorithm) should be mixtures of the first four digits.

We train in batches of 64, with a learning rate of 0.00005, on three-digit numbers in instances of size 4, 8 and 16. For size $n$, we set $a=n$. The results are shown in Figure~\ref{figure:sorting-results}.

We note that for size 4 instances, the method performs very well, reaching an error below 10\%. For size 8, the performance converges to a similar value, but with greater variance, and for size 16 we reach a minimum of around 40\% error. We also note that these experiments took relatively long to run: a single run for each size took on average 1h 40m, 7h 10m and 32h 14m respectively. 

Whether the method in its current form can usefully be applied, depends on the setting: for tasks that are easier than digit recognition, performance may converge sooner, and we may be able to sort larger numbers of items. In other settings, such as sorting by user preference for recommendation, sorting with a medium amount of accuracy may already yield acceptable results.

\section{Related work}

Learning of sparse transformations has a long history. Approaches based on parametrizing sparse transformations by dense weight vectors include the use of loss terms like L1-regularization \cite{ngiam2011sparse,han2015learning,park2018achieving} and transformations like Sparsemax \cite{martins2016softmax} and Gumbel-Sinkhorn \cite{mena2018learning} that transform the dense input to something closer to a sparse tensor. Sparsity in hidden units and other representations has also been extensively studied \cite{makhzani2013k,olshausen1997sparse} and found to be beneficial (whether our method may be effective in this area is a matter of future work).

Another way to induce sparsity in neural networks is through \emph{pruning}: enforcing sparsity, and actively setting low values to zero \cite{wen2016learning,han2015deep,han2015learning}. In this context sparsity is often employed as a way of compressing neural networks: reducing their descriptive complexity while maintaining performance.

The notion of adaptivity, making the weights of a transformation conditional on the input, or reparametrizing the weights to constrain the structure, goes back at least to \cite{schmidhuber1993self} with more recent examples including \cite{ha2016hypernetworks,krueger2017bayesian,shirakawa2018dynamic}.

Our method can also be cast as a way to learn the structure or topology of a neural network along with its weights. Most existing methods for in this area use evolutionary methods \cite{loshchilov2016cma} or reinforcement learning \cite{baker2016designing,zoph2016neural}. Approaches that do operate purely through backpropagation (apart from the hypernetworks already mentioned) include \cite{srinivas2015learning}, which introduces a trainable ReLU unit to dynamically disable nodes, and differentiable plasticity \cite{miconi2018differentiable}, which similarly uses a per-connection relevance weight, derived using Hebbian principles, to allow connections to be ignored. 
To the best of our knowledge, the sparse layer is the first that allows sparse connections to be learned without a dense parametrization.

\section{Conclusion}

We have introduced a mechanism for learning sparse transformations, using sparse datastructures. It works by defining a simple probability distribution over the  indices of the matrix. Instead of sampling one deterministic matrix, as our \texttt{REINFORCE} baseline does, we sample multiple sparse connections within the same matrix, and distribute the value over multiple connections. In other words: where the stochastic nodes of \texttt{REINFORCE} keep the topology of the computation graph fixed and sample random values, we sample a random computation graph, which consists entirely of deterministic nodes. The benefit is that the gradient which is backpropagated matches the forward computation exactly, and accurately distributes the loss.

\subsection{Limitations and future work}

In our experiments, we have shown primarily that the sparse layer can be learned effectively at a certain scale, and works when embedded in larger architectures. Further research is required to fully map out the method's potential and pitfalls.

The model seems to have a strong trade-off between the sparsity of sampled index tuples and the number of iterations required for the model to converge. In many settings, especially when the model is entirely unconstrained, sampling a sparse proportion of the total matrix may lead to infeasibly long training times. Whether a suitable tradeoff can be found depends on the task, and on whether an effective constraint for the sparse structure can be devised.

We have not yet shown that the sparse layer is effective at a scale where the matrices in question could absolutely not be embedded in a dense datastructure. Early experiments show that learning can be performed at this scale, although convergence to a very precise solution (as in the identity experiments) is difficult to attain.

While we have presented our method in the context of sparsely connected tensor contractions, this may not be the most generic formulation. Utimately, the principle applies to any situation where single, unfeasibly large computation graph can be explored by defining a probability distribution on sparse subgraphs. If we can reasonably expect such a distribution to converge to a single subgraph, we can train a model without ever seeing the whole computation graph. This view may lead to generalizations beyond sparse tensor operations, and to a more formal treatment of the method, and its properties.

\paragraph{Acknowledgements} This publication was supported by the Amsterdam Academic Alliance Data Science (AAA-DS) Program Award to the UvA and VU Universities. All experiments were executed on the DAS-5 cluster \cite{bal2016medium}; we thank the administrators for their kind assistance.

\bibliography{sparse-hyper}
\bibliographystyle{plain}

\appendix
\section{Appendix}

\subsection{Hyperparameters and other experimental details}

All experiments were executed on a single compute node, containing a single GTX TitanX GPU (Maxwell generation) with 12 GB on-board memory, dual 8-core 2.4 GHz (Intel Haswell E5-2630-v3) CPUs and 64 GB memory.

All experiments were implemented in Pytorch 0.4 \cite{paszke2017automatic}. The Adam optimizer \cite{kingma2014adam} was used in all experiments. Standard neural network layers were initialized using the Pytorch defaults. Specifically, uniform Glorot initialization for fully connected layers and convolution layers. No dropout or early stopping were used in any experiments. No weight decay, clipping or regularization was used.

\paragraph{Identity} 
A batch size of 64 is used in all experiments. For the baseline, we test the 9 learning rates $0.1, 0.05, 0.01, \ldots, 0.00001$ and report the best result. For the sparse layer, all hyperparameters are tuned by hand. In all cases this resulted in a learning rate of 0.005, except for 0.001 for the size-64 experiment. For the size $n$ of 8, 16, 32 and 64 experiments, we set $(a_l, a_g)$ to $(1,2)$, $(2,2)$, $(2,8)$ and $(2,10)$ respectively. The sampling region $(l_1, l_2)$ is set to $(\log_2 n, \log_2 n)$.

\paragraph{Attention}

\begin{table}[hbt]
  \begin{tabular}{lllll}
  type & out ch./size & kernel & padding & activation\\
  \hline
	Conv2d & 32 & 3,3 & 1 & ReLU \\
	Conv2d & 32 & 3,3 & 1 & ReLU \\
	MaxPool & & 4, 4 & 0 & \\
	Conv2d & 64 & 3,3 & 1 & ReLU \\
	Conv2d & 64 & 3,3 & 1 & ReLU \\
	MaxPool & & 2, 2 & 0 & \\
	Conv2d & 128 & 3,3 & 1 & ReLU \\
	Conv2d & 128 & 3,3 & 1 & ReLU \\
	Flatten & & & & \\
	Linear & 512 & &  & ReLU \\
	Linear & 512 & & & ReLU \\
	Linear & $4 \cdot g$ &&  &  \\
    \hline
  \end{tabular}
  \caption{Architecture of the source network for the attention experiment. $g$ is the number of glimpses used.}
  \label{table:source-arch}
\end{table}

Table~\ref{table:source-arch} shows the architecture of the source network. The glimpses are flattened, concated and fed to a 2-layer MLP with a 128-unit ReLU activated hidden layer, and a 10-unit softmax output layer. We use binary cross-entropy loss. We train for 70 epochs, using a batch size of 128 and a learning rate of 0.0001. We set $a_g=16$, $a_l=32$ and use a sampling region of $(20, 20)$.

In rare cases, the 1-glimpse model enters a state with all elements of $\D$ at high values, at which point the gradient through the sigmoid function vanishes, and training fails. Such cases can be eliminated by resetting the parameters, or re-running the model with a different random seed. This behavior has not been observed for the network with multiple glimpses.

\paragraph{Sorting}

\begin{table}[hbt]
  \begin{tabular}{lllll}
  Convnet & $(1, 28, 28)$ input & & & \\
  type & out ch./size & kernel & padding & activation\\
  \hline
	Conv2d & 16 & 3,3 & 1 & ReLU \\
	Conv2d & 16 & 3,3 & 1 & ReLU \\
	Conv2d & 16 & 3,3 & 1 & ReLU \\
    BatchNorm2d \\
	MaxPool & & 2, 2 & 0 & \\
	Conv2d & 64 & 3,3 & 1 & ReLU \\
	Conv2d & 64 & 3,3 & 1 & ReLU \\
	Conv2d & 64 & 3,3 & 1 & ReLU \\
	BatchNorm2d \\
	MaxPool & & 2, 2 & 0 & \\
	Conv2d & 128 & 3,3 & 1 & ReLU \\
	Conv2d & 128 & 3,3 & 1 & ReLU \\
	Conv2d & 128 & 3,3 & 1 & ReLU \\
	BatchNorm2d \\
	MaxPool & & 2, 2 & 0 & \\
	Flatten & & & & \\
	Linear & 256 & &  & ReLU \\
	Linear & 128 & & & ReLU \\
	Linear & 8 & &  &  \\
    \hline
	MLP & $(8\cdot d)$ input & & & \\
	\hline
	Linear & 256 & & & ReLU \\
	Linear & 1 & & & \\	
  \end{tabular}
  \caption{Architecture of the key network. Biases were disabled in layers directly preceding batch normalization. Each digit is fed through the ConvNet separately. The results are concatenated and fed through the MLP to produce a single key for the three-digit number.}
  \label{table:tokeys}
\end{table}

Table~\ref{table:tokeys} shows the architecture of the key extraction network.

\subsection{Implementation notes}

\paragraph{Flattening tensors} Sparse tensor support is limited in many frameworks, and it may not be possible to implement (\ref{eq:tensormult}) as it is described here. In such cases, the operation can be implemented by flattening $\x$ into a vector $\x'$, flattening $\W$ into a matrix $\W'$ and reshaping the resulting vector $\y' = \W'x'$ into a tensor $\y$. Here, we provide some details on how to implement such an operation. Let $\x\in \R^{n_1 \times \ldots \times n_N}$. Let $fl(\x) = \x'$ be the canonical ``flatten'' function, defined so that $\x'$ is a vector of $i_1 \times \ldots \times i_n$ elements, with 
\[
\x[a_1, ..., a_n] = \x'[fi(\{a_i\}, \{n_i\})]\,\,\,\, \text{with}\,\,\,\, fi(\{a_i\}, \{n_i\}) = \sum_i a_i \prod_{j = i+1}^N n_j\p
\]  
Let $rs(\x', (i_1, \ldots, i_n))$ be the inverse of $fl(\cdot)$. We first compute the index-tuples and values for the regular sparse tensor as before. We then translate each index tuple in $\D'$ to an index pair for $W'$. If the original index tuple is $( b_1, ..., b_m, a_1, ..., a_n)$, we translate this to $(fi(\{b_j\}, \{m_j\}), fi(\{a_i\}, \{n_i\}))$ and construct a sparse matrix $\W'$ for these indices (with the original values). We compute the vector $\y' = \W'\x'$, and reshape it to retrieve $\y$ as $\y = rs(\y', (m_1, \ldots, m_M))$.

Support for a batch dimension must also be implemented carefully: in the adaptive setting, each instance generates a different sparse matrix. In our implementation we effectively flatten a batch of $b$ sparse $n \times m$ matrices into a single larger $bn \times bm$ matrix with the matrices of the batch arranged along the diagonal. This allows us to compute the batch using a single sparse matrix multiplication.

Note that the translation from real to integer indices happens in the original tensor rank. The flattening of the index tuples only happens once the integer tuples have been computed.

\paragraph{Gradient accumulation} In a naive implementation, distributing the probabilities of all continuous tuples over all sampled integer tuples and normalizing requires a dense matrix that is likely larger than $\W$. If $\W$ is large enough to make this infeasible, we compute the gradient in multiple passes. For each pass, we compute the gradient for only a subset of the continuous tuples, in a complete forward and backward pass. The computation graph is then cleared and the process is repeated for the next subset of continuous tuples (keeping the sample of integer tuples the same), accumulating the gradients over the parameters.

\paragraph{Detecting duplicates} For efficient computation of the sparse layer, we have found that the best approach is to sample the integer index tuples with replacement, uniformly over a region, and to set the probability of duplicate tuples to zero. This ensures that tuples do not receive extra probability mass from being sampled twice, yet allows us to sample with replacement, with is more efficient.

For a given matrix $\D' \in \R^{k\times r}$ of sampled integer tuples, we convert each row $(d_1, \ldots, d_r)$ to a unique integer using the Cantor tuple function:
\begin{align*}
u(d_1, d_2) &= \frac{1}{2}(d_1 + d_2)(d_1 + d_2 + 1) + d_2 \\
u(d_1, \ldots, d_r) &= u\left(d_1, u(d_2, \ldots, d_r)\right)
\end{align*}
We then (non-differentiably) sort the resulting vector of unique indices, $\mathbold{u}$, and compare it to a vector $\mathbold{u}'$ defined as $\mathbold{u}'[i] =\mathbold{u}'[i+1]$, to create a binary vector that masks all repeated occurrences of the same tuple (but not the first). We reverse the sort to apply the mask to the original tuples.
 
Finally, we use this mask to set the probabilities corresponding to these tuples to zero. It is important to do this \emph{before} normalizing the probabilities, or they will not sum to $1$.

\paragraph{Computing $P^d(\bo)$ from $\bo$}

$P^d(\bo) \in \R^{n\times n}$is a permutation matrix: it has a unique nonzero value in each column and in each row, each taking value 1. We will compute a sequence $\br(\bo)$ such that the index tuples $(\br(\bo)_i, i)$ with $i \in [1..n]$ represent the nonzero elements of $P^d(\bo)$. Algorithm~\ref{algorithm:compute-hperm} computes this sequence from $\bo$.

\begin{pseudo}[h]
\caption{Compute $\br(\bo)$ from $\bo$. The function $\text{cumsum}(x, c)$ computes the cumulative sum over $x$, resetting after every $c$ elements.}
\label{algorithm:compute-hperm}
~\\
$\bu = [1..2^d] \cdot n2^{-d} +  n2^{-(d+1)}$ \hfill{\# starting indices of upper buckets} \\
$\bl = [1..2^d] \cdot n2^{-d}$ \hfill \# starting indices of lower buckets\\
~\\
$\bc_\bu \leftarrow \text{cumsum}(\bo, n2^{-d})$\\
$\bc_\bl \leftarrow \text{cumsum}(1-\bo, n2^{-d})$\\
~\\
\textbf{return} $\br \leftarrow \bo \cdot (\bc_\bu - 1) + (\bo-1) \cdot (\bc_\bl - 1)$\\
\end{pseudo}

\end{document}